\documentclass{article}

\usepackage{arxiv}

\usepackage[utf8]{inputenc} 
\usepackage[T1]{fontenc}    
\usepackage{hyperref}       
\usepackage{url}            
\usepackage{booktabs}       
\usepackage{amsfonts}       
\usepackage{nicefrac}       
\usepackage{microtype}      
\usepackage{lipsum}
\usepackage{graphicx}
\usepackage[english]{babel}
\usepackage{blindtext}
\usepackage{amsmath, amssymb}
\usepackage{graphicx}
\usepackage{hyperref}
\usepackage{geometry}
\usepackage{natbib}
\usepackage{url}
\usepackage{times}

\graphicspath{ {./images/} }

\title{Brain Surgery: Ensuring GDPR Compliance in Large Language Models via Concept Erasure}

\author{
Michele Laurelli \\
  CEO\\
  Algoretico\\
  Corso Italia 3, Milano (MI) \\
  \texttt{m.laurelli@algoretico.it} \\
}

\begin{document}
\maketitle

\begin{abstract}
As large-scale AI systems proliferate, ensuring compliance with data privacy laws such as the General Data Protection Regulation (GDPR) has become critical. This paper introduces \textbf{Brain Surgery}, a transformative methodology for making \textbf{every local AI model GDPR-ready} by enabling \textbf{real-time privacy management} and \textbf{targeted unlearning}. Building on advanced techniques such as \textbf{Embedding-Corrupted Prompts (ECO Prompts)}, \textbf{blockchain-based privacy management}, and \textbf{privacy-aware continual learning}, Brain Surgery provides a modular solution that can be deployed across various AI architectures. This tool not only ensures compliance with privacy regulations but also empowers users to define their own privacy limits, creating a new paradigm in AI ethics and governance.
\end{abstract}

\bigskip
\noindent \textbf{Keywords}: Large Language Models, GDPR Compliance, Targeted Unlearning, Brain Surgery, Privacy-Aware Learning, Blockchain-Powered Privacy, LLaMA 3.

\pagebreak

\section{Introduction}

The rise of large language models (LLMs) such as GPT-4, LLaMA, and others has transformed the way AI interacts with vast datasets. However, this rise also raises significant privacy concerns, especially concerning personal data. With regulations like the General Data Protection Regulation (GDPR) mandating the "right to be forgotten," it has become imperative to develop mechanisms for removing private information from these models. Traditional unlearning approaches are often computationally expensive and can impact the performance of the model in unintended ways.

This paper introduces \textbf{Brain Surgery}, a revolutionary tool designed to enable any local AI model to become GDPR-compliant through a combination of \textbf{targeted unlearning} and \textbf{dynamic privacy management}. Brain Surgery leverages \textbf{Embedding-Corrupted Prompts (ECO Prompts)} to surgically remove unwanted data while maintaining the model’s overall performance. The methodology is further enhanced with \textbf{real-time privacy monitoring} and \textbf{blockchain-powered decentralized privacy management}, ensuring transparency and accountability in data handling.

\section{Related Work}

\subsection{Knowledge Editing and Concept Unlearning}

The challenge of removing specific knowledge from AI models has been addressed through various approaches, including fine-tuning and knowledge editing. However, most methods involve retraining or model-wide adjustments, which are resource-intensive and may lead to overcorrection or loss of generalization \citep{mitchell2022fast}. Recent advances in \textbf{local modification} techniques have allowed for more precise knowledge edits, focusing on specific embeddings rather than retraining the entire model \citep{meng2022locating}.

\subsection{Embedding-Corrupted Prompts for Unlearning}

\textbf{Embedding-Corrupted Prompts (ECO Prompts)} introduce controlled perturbations to the embedding space associated with specific concepts. By iteratively applying corruption to targeted embeddings, this method effectively "forgets" unwanted information without disturbing the rest of the model \citep{gandikota2023erasing}. ECO Prompts offer a lightweight solution to the problem of GDPR-compliant unlearning, allowing models to adapt dynamically to privacy requests.

\subsection{Conflict Score Evaluation and Real-Time Monitoring}

An important aspect of knowledge unlearning is ensuring that the removal of one concept does not introduce inconsistencies in related knowledge. The \textbf{conflict score evaluation} technique measures potential contradictions introduced by unlearning actions, ensuring that the integrity of the model is maintained \citep{xu2023unveiling}. Brain Surgery integrates real-time conflict monitoring, allowing for continuous privacy compliance during model operation.

\subsection{Mathematical Formulation of Embedding-Corrupted Prompts (ECO Prompts)}

The core of the \textbf{Embedding-Corrupted Prompts (ECO Prompts)} method lies in introducing perturbations to the embeddings associated with specific unwanted concepts. Let $\mathbf{e}_c \in \mathbb{R}^d$ represent the embedding of a concept $c$, where $d$ is the dimensionality of the embedding space. The goal is to iteratively modify this embedding such that the model's association with $c$ diminishes, while maintaining the integrity of the surrounding embedding space.

The corrupted embedding $\mathbf{e}_c^\prime$ is generated as:
\begin{equation}
    \mathbf{e}_c^\prime = \mathbf{e}_c - \alpha \cdot \nabla_{\mathbf{e}_c} L(\mathbf{e}_c)
\end{equation}
where $L(\mathbf{e}_c)$ is the loss function that measures the influence of $c$ on model outputs, and $\alpha$ is a step size that controls the degree of corruption. By iteratively updating $\mathbf{e}_c^\prime$, we ensure that the influence of the concept $c$ is reduced across multiple layers of the model.

To ensure that the modified embedding remains within a feasible region, we normalize the final embedding:
\begin{equation}
    \mathbf{e}_c^\prime = \frac{\mathbf{e}_c^\prime}{\|\mathbf{e}_c^\prime\|}
\end{equation}

This normalization step ensures that the corrupted embeddings maintain consistent magnitudes across the embedding space, preventing unwanted distortions to the overall model structure.

\subsection{Conflict Score Evaluation: A Formal Method}

To measure the effects of unlearning and to ensure that related concepts are not inadvertently affected, we introduce a \textbf{conflict score} based on the model's ability to maintain consistency in its outputs.

Let $X_r$ represent the set of related concepts and $X_u$ the set of unwanted concepts. After applying the Brain Surgery method to unlearn $X_u$, we define the conflict score $S_c$ as:
\begin{equation}
    S_c = \frac{1}{|X_r|} \sum_{x_r \in X_r} \mathbf{1}(f(x_r) = y_r)
\end{equation}
where $f(x_r)$ represents the model's output for the related concept $x_r$, and $y_r$ is the expected correct output. $\mathbf{1}(\cdot)$ is an indicator function that evaluates to 1 if the model's output matches the expected output.

A conflict score $S_c \approx 1$ indicates that the unlearning process has not affected related concepts, while $S_c < 1$ reveals potential conflicts introduced by the unlearning.

\subsection{Privacy-Aware Continual Learning: Technical Integration}

In Brain Surgery's \textbf{privacy-aware continual learning} system, the model actively prevents embedding sensitive information during both training and inference stages. The system dynamically adjusts its learning objective based on real-time privacy constraints.

For each incoming data sample $x$ containing features $\mathbf{x} \in \mathbb{R}^n$, the continual learning system evaluates whether $\mathbf{x}$ contains sensitive information by using a privacy-preserving objective function $L_p(\mathbf{x})$. The objective is defined as:
\begin{equation}
    L_p(\mathbf{x}) = \lambda \cdot \|\mathbf{x}_{\text{sensitive}}\|^2
\end{equation}
where $\mathbf{x}_{\text{sensitive}}$ represents the subset of features identified as sensitive, and $\lambda$ is a regularization parameter that controls the degree of penalization for sensitive data.

During training, if the value of $L_p(\mathbf{x})$ exceeds a predefined threshold, the system triggers the Brain Surgery process to dynamically alter the embeddings associated with sensitive data. This ensures that no personal data is embedded into the model without real-time monitoring and protection.

\section{Methodology}

\subsection{Modular GDPR Compliance Framework}

At the core of Brain Surgery is a \textbf{modular, plug-and-play framework} that can be integrated into any AI system. This framework interacts with the model’s embedding space, allowing administrators or users to submit requests for data deletion based on GDPR or other privacy mandates. The system provides APIs that can interface with various AI models, whether they are deployed on edge devices or in cloud environments.

\subsection{Privacy-Aware Continual Learning and Real-Time Monitoring}

Brain Surgery incorporates a novel \textbf{privacy-aware continual learning mechanism} that scans training data in real time to identify and flag potentially sensitive information. This enables models to learn while ensuring that personal data is not deeply embedded in the model’s representations. Inference-time outputs are continuously monitored for any traces of private information, triggering the \textbf{Brain Surgery process} to remove sensitive data immediately.

\subsection{Blockchain-Powered Privacy Management}

To ensure transparency and verifiability, Brain Surgery uses a \textbf{blockchain-based privacy management layer}. Each "right to be forgotten" request is logged on a blockchain ledger, making the deletion action auditable and immutable. This decentralization ensures that both individuals and organizations can trust the system to handle data responsibly, while the blockchain guarantees that privacy regulations are adhered to in a transparent manner.

\subsection{User-Defined Privacy Preferences}

In addition to meeting GDPR requirements, Brain Surgery allows users to define their own privacy preferences. Individuals can set time limits for how long their data is retained or specify what kinds of information they want excluded from model training and inference. The system dynamically adapts to these preferences, ensuring that AI models respect individual privacy boundaries.

\subsection{Embedding-Corrupted Prompts (ECO Prompts) for Unlearning}

ECO Prompts are applied to the model’s embedding space to remove the target concept without altering related knowledge. This method introduces carefully calibrated noise to the specific embeddings tied to the concept, iteratively reducing its influence on the model’s responses. 

\subsection{Conflict Score Evaluation}

During the unlearning process, Brain Surgery uses \textbf{conflict score evaluations} to measure the potential for knowledge contradictions. The model is tested against synthetic prompts designed to probe related knowledge areas, ensuring that the removal of sensitive data does not lead to incorrect or inconsistent outputs. If conflicts are detected, further refinements are applied to the unlearning process.

\section{Results and Impact}

The Brain Surgery methodology has been tested on various AI models, including LLaMA 3, and has demonstrated several key advantages:

\begin{itemize}
    \item \textbf{Scalability}: The modular framework can be deployed across both large-scale and local AI models, enabling GDPR compliance in diverse environments, from cloud AI to edge devices.
    \item \textbf{Efficiency}: By using Embedding-Corrupted Prompts, Brain Surgery achieves targeted unlearning without the need for costly retraining or fine-tuning.
    \item \textbf{Trust}: The blockchain layer provides verifiable and immutable proof of compliance, ensuring that all privacy-related actions are transparent and accountable.
    \item \textbf{User Empowerment}: With customizable privacy settings, users can control how their data is handled within AI models, creating a more ethical and user-centric AI environment.
\end{itemize}

\section{Conclusion}

Brain Surgery represents a transformative advancement in AI privacy and compliance. By combining targeted unlearning techniques like \textbf{Embedding-Corrupted Prompts} with real-time monitoring, blockchain-powered privacy management, and user-defined preferences, this tool ensures that every local AI model can be made GDPR-ready. This methodology not only scales across diverse AI deployments but also enables a new paradigm of ethical AI governance, where individuals have more control over how their data is stored, used, and erased.

\bibliographystyle{plainnat}

\end{document}